\documentclass[10pt,twocolumn]{article}
\pdfoutput=1
\usepackage{times}
\usepackage{graphicx}
\usepackage{amssymb}
\usepackage[ruled,vlined,noend]{algorithm2e}
\usepackage[utf8]{inputenc} 
\usepackage[T1]{fontenc}    
\usepackage{hyperref}       
\usepackage{url}            
\usepackage{booktabs}       
\usepackage{amsfonts}       
\usepackage{nicefrac}       
\usepackage{microtype}      
\usepackage{lipsum}
\usepackage{booktabs}
\usepackage{multirow}
\usepackage{setspace}
\usepackage{makecell}
\usepackage{float}
\usepackage{multicol}

\newtheorem{definition}{Definition}[section]

\begin{document}

\date{}

\title{One vs Previous and Similar Classes Learning - A Comparative Study}

\author{
  Daniel Cauchi \\
  Faculty of Information \\
  \& Communication Technology\\
  University of Malta\\
  Msida, MSD 2080 \\
  \texttt{daniel.cauchi.17@um.edu.mt} \\
  \and
  Adrian Muscat \\
  Faculty of Information \\
  \& Communication Technology\\
  University of Malta\\
  Msida, MSD 2080 \\
  \texttt{adrian.muscat@um.edu.mt}
}
\twocolumn[
\begin{@twocolumnfalse}
\maketitle
\begin{abstract}
When dealing with multi-class classification problems, it is common practice to build a model consisting of a series of binary classifiers using a learning paradigm which dictates how the classifiers are built and combined to discriminate between the individual classes. As new data enters the system and the model needs updating, these models would often need to be retrained from scratch. This work proposes three learning paradigms which allow trained models to be updated without the need of retraining from scratch. A comparative analysis is performed to evaluate them against a baseline. Results show that the proposed paradigms are faster than the baseline at updating, with two of them being faster at training from scratch as well, especially on larger datasets, while retaining a comparable classification performance.
\end{abstract}

\end{@twocolumnfalse}
]

\section{Introduction}
For multi-class classification problems~\cite{Multiclass_Classification}, models are often built using a series of binary classifiers. The model uses a protocol to build the binary classifiers during training so it is then able to perform prediction using the built classifiers and the previously mentioned protocol. In this work, this protocol is referred to as the \textit{learning paradigm}, or just \textit{paradigm}. Examples of existing paradigms are One vs Rest (OvR)~\cite{In_Defence_of_OvR} and One vs One (OvO), also known as Round Robin Rule Learning~\cite{Round_Robin_Rule_Learning, Multiclass_Classification}. This work proposes three paradigms. The first, \textit{Similar Classes Learning} (SCL), is a modified version of the paradigm proposed by~\cite{Learning_Cumulatively_to_Become_More_Knowledgeable}. \textit{One vs Previous} (OvP) is a new paradigm and \textit{One vs Previous Similar Classes} (OvPSC) is a combination of the previous two.

Existing paradigms are usually highly computationally demanding~\cite{OvR_ComputationallyExpensive, OvR_ComputationallyExpensive2} and if a model built with these paradigms needs to be updated with new data, it often needs to be retrained from scratch. Each of the proposed paradigms decouple the classifiers of the model in some way, such that, as new data enters the system, only relevant parts of the model need to be modified. This gives them application in Lifelong Machine Learning~\cite{Lifelong_Machine_Learning_Systems, Lifelong_Machine_Learning:_a_paradigm_for_continuous_learning}.

\begin{figure*}
  \centering
  \includegraphics[width=\textwidth]{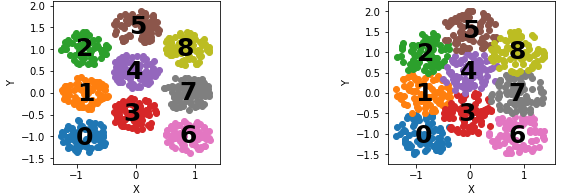}
  \caption{The synthetic clusters datasets with radius=0.4 (left) and radius=0.525 (right). Each number corresponds to a class label.}\label{fig:CirclesDatasetVaryingRadius}
\end{figure*}

Two datasets are used to evaluate the paradigms against a baseline. The first is a real-world dataset of spatial relations in images, SpatialVOC2K~\cite{spatialvoc2k, Third_Dimension_to_2D_images, GabrielFarrugia}. The second is a synthetic dataset consisting of nine clusters, which can be seen in Figure~\ref{fig:CirclesDatasetVaryingRadius}, used to experiment freely with its characteristics.

The contribution of this paper are the following: a) propose the three paradigms, one of which is a modification of a known method, one is new and one which is a combination of the previous two b) conduct a comparative analysis against a bassline and evaluate them on their computational time and classification performance, using two datasets, one of which is synthetic and the other is from a real world scenario.

\section{Related Work}
This section provides an overview of contributions related to this work and how the proposed paradigms differ from the existing alternatives.

One of the standard paradigms is One vs Rest (OvR)~\cite{In_Defence_of_OvR}. Here, each class is trained against all other classes resulting in as many classifiers as classes. For prediction, an argmax function is used to determine the output class. The nature of this paradigm is simple but is of a brute force nature. While the proposed paradigms share some similarities with this paradigm, they also differ in such a way that simplicity is traded for efficiency. Certain other paradigms claim to be have better classification performance, however the evidence suggesting this is usually not sufficient or results may be biased~\cite{In_Defence_of_OvR}. Hence why OvR remains commonplace and is the reason it is used as a baseline paradigm for this work.

A standard alternative to OvR is One vs One (OvO)~\cite{Multiclass_Classification, Round_Robin_Rule_Learning}, where every class is trained every other class in pairs, resulting in a model characterized by $\frac{n(n-1)}{2}$ classifiers. In the prediction phase, a scoring method is used and the class with the most winning classifiers is chosen. The prediction algorithm treats the classifier output as binary, without considering the score. This is unlike OvR which compares the score of each classifier to determine the output. Both OvP and OvPSC treat the classifier output as binary as well.

\cite{CLEAR} uses a one class SVM for each class, allowing the classifiers for each class to be decoupled from the data of other classes. This allows the model to mitigate disruption when new classes are introduced; a concept called \textit{Cumulative Learning}, which is a sub-area of Lifelong Machine Learning~\cite{Lifelong_Machine_Learning_Systems}.

Cumulative learning is also used by~\cite{Learning_Cumulatively_to_Become_More_Knowledgeable}. With this method, as a new class comes in, it is trained against its \textit{similar} classes only, while the classifiers for the \textit{similar} classes are retrained to include the new class in their negative set. \cite{Learning_Cumulatively_to_Become_More_Knowledgeable}. SCL uses a similar algorithm, however, significant additions were made to generalize the model such that it can be used by any multi-class model that uses a series of binary classifiers. On the other hand, \cite{Learning_Cumulatively_to_Become_More_Knowledgeable} solves the problem of Open World Classification \cite{Breaking_The_Closed_World_Assumption}, where the classes in the test set are not assumed to have all been seen during training.

\section{The Paradigms} \label{sec:TheParadigms}
This section is dedicated to discussing the individual paradigms. Each paradigm consists of two algorithms: the training algorithm and the prediction algorithm. Before starting with the algorithms themselves, the term \textit{similarity}, and by extension \textit{similar classes}, is defined, which are used by SCL and OvPSC. The definition provided by~\cite{Learning_Cumulatively_to_Become_More_Knowledgeable} is used:
\begin{definition}

	\textbf{Similar Classes (SC)}: Given an existing model at iteration $t$ with classifiers ${C_1, C_2, ..., C_t}$ for classes ${c_1, c_2, ..., c_t}$ respectively, when at iteration $t+1$, the $N_{t+1}$ examples for class $c_{t+1}$ enter the system, these are run through each classifier of the past classes and if a percentage $\lambda_{sim}$ of these examples are classified as positive for a classifier, then the new class and the class corresponding to that classifier are said to be similar.
\end{definition}

$\lambda_{sim}$ is a hyperparameter. For the experiments, this was set to a constant 2\%, which was the value used in~\cite{Learning_Cumulatively_to_Become_More_Knowledgeable} and through preliminary testing, this value was also found to work best for the used datasets. If it is taken to be lower, then too many classes are marked as similar, reducing the efficiency of the paradigms and by extension the models created by them. If it is taken to be higher, then not enough classes are marked as similar, leading to a decline in classification performance.

With regards to algorithm notation, data types are underlined, keywords are in bold text, \# in front of a list or set represents its size, lists are ordered and are represented within [] while sets are unordered and are represented within \{\}.

  \subsection{Similar Classes Learning (SCL)}
This paradigm is a modification on~\cite{Learning_Cumulatively_to_Become_More_Knowledgeable}. The modifications allow it to solve a standard multi-class classification problem.

\begin{algorithm}[h]
	 \KwIn{\underline{Sets of Training Examples} $E_1, E_2, \ldots{}, E_m$ for \underline{Classes} $c_1, c_2, \ldots{}, c_m$ \newline 
		  \underline{Base Classifier} B \newline
		  \underline{float} $\lambda_{sim}$
	 }
	 \underline{Model} M = [ ]
	
	 \ForEach{\underline{int} i \textbf{in} [1\ldots{}m]}{
		\underline{Set of classes} $SC_i = \{\}$
		
		\underline{Set of classes} $CannotBe_i = \{\}$
		
		\ForEach{\underline{Classifier} $C_j$ \textbf{in} M}{
			\underline{int} count = 0
			
			\ForEach{\underline{Training Example} e \textbf{in} $E_i$}{
				\If{$C_j$ classifies e as positive}{
					count = count + 1
				}
			}
			\eIf{count $> \lambda_{sim} * \#E_i$}{
				$SC_i = SC_i \cup c_j$
				
				$SC_j = SC_j \cup c_i$
				
				Retrain $C_j$ with $c_j$ as the positive class and $SC_j$ as the negative classes
			}{
				$CannotBe_j = CannotBe_j \cup c_i$
			}
		}
		\eIf{SC is empty}{
			$C_i$ = \textit{DummyClassifier}
		}{
			Train $C_i$ of type B with $c_i$ as the positive class and $SC_i$ as the negative classes
		}
		\textbf{Append} $C_i$ \textbf{to} M
		
	 }
	 \caption{SCL Training}\label{algo:SCL-training}
\end{algorithm}

When training (Algorithm \ref{algo:SCL-training}), the inputs are the training examples for each class and a base classifier. These inputs are the same as for that of OvR. The base classifier can be any binary classifier, such as a logistic regressor or support vector machine. The choice of classifier will affect how the decision boundaries are built. The last input is $\lambda_{sim}$ discussed previously.

Starting from an empty model, the algorithm iterates over each class, initializing an empty Similar Classes (\textit{SC}) set and CannotBe set (lines 1 - 4). The reasoning behind \textit{CannotBe} is: if all the examples of a class B, save for a few outliers, have been classified as negative by a classifier for a class A, then if another example is introduced to the system and the classifier for class A has classified it as positive, then it cannot be, or at least it is highly unlikely, that it belongs to class B. This \textit{CannotBe} variable is what distinguishes this work from~\cite{Learning_Cumulatively_to_Become_More_Knowledgeable} and is what allows the paradigm to be used for the Closed World Assumption~\cite{Breaking_The_Closed_World_Assumption}, where every class in the testing data is assumed to have been seen during training. Using the examples of the current class, the algorithm finds the \textit{SC} of this class using existing classifiers (lines 5 - 10), adds each class to the \textit{SC} of the other class symmetrically (lines 11-12) and retrains the classifier of the existing class with its previous negative examples and the new class as the negative set (line 13). Otherwise the current class is added to the \textit{CannotBe} set of the existing class (lines 14 - 15). If a class is not similar to any other class, it is given a dummy classifier which gives the highest positive score to any example during prediction (lines 16 - 17). In contrast, if similar classes are found, the class is given a classifier which is trained with its similar classes as the negative set (lines 18 - 19). Lastly, this new classifier is appended to model (line 20).

\begin{algorithm}[h]
	 \KwIn{\underline{Example} to be classified e \newline
		  \underline{Model} M = [ $C1, C2, \ldots{}, C_m$ ] \newline
		  \underline{List} of \underline{set of classes} $[CannotBe_1, CannotBe_2, \ldots{}, CannotBe_m]$
	 }

	 TopClass = NULL
  
	 BiggestConfidence = -$\inf$
	
	 CannotBe = \{\}

	\ForEach{\underline{int} i \textbf{in} [1\ldots m]}{
 	\If{$c_i$ \textbf{not in} CannotBe}{
      \underline{float} conf = Confidence in e given by $C_i$
      
      \If{conf \textgreater{} BiggestConfidence}{
        BiggestConfidence = conf
      
        TopClass = $c_i$
      }

      \If{conf \textgreater{} 0}{
        $CannotBe = CannotBe \cup CannotBe_i$
       }
    }
 	}
 	\textbf{return} TopClass
	\caption{SCL Prediction}
	\label{algo:SCL-prediction}
\end{algorithm}

As inputs for prediction (Algorithm~\ref{algo:SCL-prediction}), besides the example to be predicted, the algorithm needs to know the \textit{CannotBe} set of each class. A local \textit{CannotBe} set is kept (line 3). The algorithm iterates over all the binary classifiers of the model in the order in which they were built, skipping over ones which are in the local \textit{CannotBe} set (lines 4 - 5). If a class gives a positive confidence score, the local \textit{CannotBe} set is unionized with the \textit{CannotBe} of the class (lines 10 - 11). Finally, it predicts the class whose classifier gives the highest confidence (lines 7 - 9, 12).

One may notice that the order in which the classes enter the system matters. A different order will lead to a different model and may affect classification performance. This observation applies to all three of the proposed paradigms and finding the optimal order is still an open question. Besides training the model, SCL and OvPSC also need to spend time searching for the similar classes during training, hence, training time also depends on the prediction speed of the base classifier used. In all the proposed paradigms, if new examples belonging to a new class come in, there is no need to retrain the entire model from scratch, similar to OvO, but unlike OvR. Unlike OvO, however, there is no need to train as many new classifiers as there are classes, but only train a single new classifier. In the worst case, when all classes are similar, the decision boundary of this paradigm will be identical to that of OvR.

  \subsection{One vs Previous (OvP)}
The next paradigm, One vs Previous (OvP), is new. In short, this starts from a dummy classifier and proceeds to make each subsequent classifier more and more specific.

\begin{algorithm}[h]
	 \KwIn{\underline{Sets of Training Examples} $E_1, E_2, ... E_m$ for \underline{Classes} $c_1, c_2, ... c_m$ \newline 
	 \underline{Base Classifier}}

	 \underline{Model} M = [ ]

	 \ForEach{\underline{int} i in [1...m]}{

		\eIf{i = 1}{
			 \underline{Classifier} $C_i$ = \textit{DummyClassifier}
		}{
     Train $C_i$ of type B with $c_i$ as the positive class and $c_1 ... c_{i-1}$ as the negative classes
		}
    \textbf{Append} $C_i$ \textbf{to} M
	 }

	 \caption{OvP Training}\label{algo:OvP-training}
\end{algorithm}

The inputs for the training algorithm (Algorithm~\ref{algo:OvP-training}) consist of the examples and base classifier. A dummy classifier is created for the first class (line 4). Subsequent classes are given a classifier which is trained using the examples of classes which already have a classifier built for them (the previous classes) as the negative examples (lines 5 - 6).

\begin{algorithm}[h]
	 \KwIn{\underline{Example} to be classified e \newline
		\underline{Model} M = [ $C1, C2, ..., C_m$ ]}

	 \ForEach{\underline{int} i in [m...1]}{
	  	\underline{float} conf = Confidence in e given by $C_i$
		
	  	\If{conf \textgreater 0}{
	   		\textbf{return} $c_i$
	  	}
	 }

 	\caption{OvP Prediction}\label{algo:OvP-Prediction}
\end{algorithm}

The prediction algorithm (Algorithm~\ref{algo:OvP-Prediction}), iterates through the classifiers in the reverse order of which they were built and predicts the first class whose classifier gives a positive confidence.

When training, the classifiers do not depend on each other. Thus, similar to OvR but unlike SCL and OvPSC, training the individual classifiers can be parallelized.

  \subsection{One vs Previous Similar Classes (OvPSC)}
This final paradigm is a combination of the previous two, where it builds the classifiers similar to SCL, without retraining, and then during prediction it uses a local \textit{CannotBe} to see which classes it should not consider, then performs an OvP prediction over the remaining classes.

\begin{algorithm}[h]
 	\KwIn{\underline{Sets of Training Examples} $E_1, E_2, ... E_m$ for \underline{Classes} $c_1, c_2, ... c_m$ \newline 
  		\underline{Base Classifier} B \newline
  		\underline{float} $\lambda_{sim}$
 	}
 	\underline{Model} M = [ ]

 	\ForEach{\underline{int} i \textbf{in} [1...m]}{
  		\underline{Set of classes} $SC_i = \{\}$
		
  		\underline{Set of classes} $CannotBe_i = \{\}$
		
  		\eIf{i == 1}{
   			\underline{Classifier} $C_i$ = \textit{DummyClassifier}
  		}{
   			\ForEach{$C_j$ \textbf{in} M}{
        \underline{int} count = 0
      
        \ForEach{\underline{Training Example} e \textbf{in} $E_i$}{
          \If{$C_j$ classifies e as positive}{
            count = count + 1
          }
        }
        \eIf{count $> \lambda_{sim} * \#E_i$}{
          $SC_i = SC_i \cup c_j$
        }
        {
          $CannotBe_j = CannotBe_j \cup c_i$
        }
      }
      Train $C_i$ of type B with $c_i$ as the positive class and $SC_i$ as the negative classes
    }
    \textbf{Append} $C_i$ \textbf{to} M
  }
  \caption{OvPSC Training}\label{algo:OvPSC-training}
\end{algorithm}

The training algorithm (Algorithm~\ref{algo:OvPSC-training}) has the same inputs as SBL. Starting from an empty model, the algorithm iterates over all the classes, giving each one its own \textit{SC} and \textit{CannotBe} set (lines 1 - 4). The first class is given a dummy classifier (lines 5 - 6). Iterating through existing classifiers, the classes which are similar to the class in consideration are found and added to the \textit{SC} of this class, however, as opposed to SBL, the relationship is not symmetric (lines 7 - 14). If the current class is not similar to the existing class, it is added to the \textit{CannotBe} of the existing class (lines 15 - 16). Lastly, a classifier is trained for the current class against its similar classes, without any modification of existing classifiers, which is then appended to the model (lines 17 - 18).

\begin{algorithm}[h]
	 \KwIn{\underline{Example} to be classified e \newline
		  \underline{Model} M = [ $C1, C2, ..., C_m$ ] \newline
		  \underline{List} of \underline{set of classes} $[CannotBe_1, CannotBe_2, ..., CannotBe_m]$
	 }

 	CannotBe = \{\}

 	\underline{Class} toReturn = NULL

 	\ForEach{\underline{int} i \textbf{in} [1...m]}{
  		\If{$c_i$ \textbf{not in} CannotBe}{
		  	 \underline{float} conf = Confidence in e given by $C_i$
			
		   	\If{conf \textgreater 0}{
				    $CannotBe = CannotBe \cup CannotBe_i$
				
				    toReturn = $c_i$
			   }
		  }
  }

  \textbf{return} toReturn

  \caption{OvPSC Prediction}\label{algo:OvPSC-Prediction}
\end{algorithm}

For prediction, (Algorithm~\ref{algo:OvPSC-Prediction}, the algorithm keeps a local \textit{CannotBe} and iterates over the classes in the order of which the classifiers were built (lines 1 - 3). If a class is in the local \textit{CannotBe}, it is skipped (line 4). Otherwise, if the classifier returns a positive confidence, the \textit{CannotBe} of that class is unionized with the local \textit{CannotBe} and the algorithm will return the current class if no other class gives a positive confidence (lines 5 - 9).

One may notice that since the first class is given a dummy classifier, it will always be similar to every other class. In the worst case, when all classes are similar, the decision boundary of this paradigm will be identical to that of OvP.

\section{Datasets}
\subsection{SpatialVOC2K}
The first dataset used is one containing spatial relations in images \cite{spatialvoc2k, Third_Dimension_to_2D_images, GabrielFarrugia}, consisting of 20 different objects and 17 different target values. It is split into two parts, of which the entitled \textit{best} part is used, containing 5317 examples, further subdivided into five stratified folds. Each feature vector contains language and geometric features. From preliminary testing and manual feature selection, the one-hot vector for the language features was chosen and the initial 34 geometric features, which extend to 37 once the feature \textit{rel\_position} is expanded into a one-hot vector, were reduced to 23 features. For the full list of features and additional visualisations of this dataset, see Appendix \ref{App:SpatRelFeatures}.

\subsection{Synthetic Clusters}
The second dataset consists of 9 circular clusters on a 2D plane of up to 100 points each as can be seen in Figure~\ref{fig:CirclesDatasetVaryingRadius}. This dataset was created for free experimentation with its characteristics. Four variants of this dataset were created, each with a varying radius: \textit{clusters\_veryFar} has radius 0.1, \textit{clusters\_far} has radius 0.4, \textit{clusters\_close} has radius 0.5 and \textit{clusters\_intersecting} has radius 0.525. At radius 0.5, the clusters are touching. For additional details and visualisations about this dataset see Appendix \ref{App:Clusters}.

\section{Experiments}
A numer of experiments were devised to show how the paradigms perform in different scenarios. With regards to base classifiers, a logistic regressor with the \textit{liblinear} solver \cite{liblinear} and an SVM with RBF Kernel \cite{Vapnik, A_Primer_on_Kernel_Methods} were used. 

The training pipeline is as follows. First, the datasets are split into five stratified folds. The non-one-hot features of this dataset are also normalized to mean 0 and standard deviation 1. The parameters for normalization are obtained on the train set and then applied to the test set. There are thus five runs, one for each fold. Before training in each run, a grid search is performed over the C parameter of both classifiers, where C ranges from 0.0001 to 100 using a logarithmic scale. Lastly, the best classifier is trained over the entire train set and evaluated over the test set. In each experiment, the classes enter the system in order of frequency, starting from the most frequent first, because in a practical scenario this would usually be the order in which classes would appear.

  \subsection{Experiment 1}
In the first three experiments, 1a, 1b and 1c, after splitting into the test and train sets, the train set is further split into three stratified parts which are called M0, M1 and M2. The first M-set, M0, contains 50\% of the data, M1 contains 30\% of the data and M2 contains the last 20\%. These proportions were chosen such that they represent a practical scenario, starting with most data at the start, then slowly decreasing. The experiment then simulates what happens when each M-set enters the system one after the other, M0 to M2, and the models are updated in accordance with the new data. The strategy for updating is what varies between the experiments. The hyperparameters for these experiments are tuned only once on the first M set, then kept constant for all the classifiers till the end of the run, as this was observed to work best for these datasets. Lastly, for experiments 1b and 1c, with regards to the SCL and OvPSC paradigms, once similar classes are determined using the first M-set, they are kept the same for the subsequent M-sets.

\subsubsection{Experiment 1a - Retraining from scratch}
For this first experiment, the model is retrained from scratch with the accumulated data as each M-set enters the system. This means that in the first iteration, it is trained with the data in M0, then when M1 enters, the weights are reset and the model is retrained with the data in both M0 and M1. The same happens for the final M-set.

\subsubsection{Experiment 1b - Fine-tune with new set}
In this experiment, when the second and third set come in, the weights are retained from the previous iteration and the model is trained until convergence using the new set only.

\subsubsection{Experiment 1c - Fine-tune with entire set}
This last experiment entails retaining the weights from previous iterations when a new set comes in and retraining with the data in both the previous and new M-set until convergence. This transferring of weights for updating used in both experiment 1b and experiment 1c is often called transfer learning \cite{Transfer_Learning}.

  \subsection{Experiment 2}
The second experiment evaluates the models when the number of classes \textit{N} changes. The train set is first split into N sets, where each set contains all the examples of a single class. Hyperparameter tuning is done each time a new class enters the system, and any classifiers which need to be trained or retrained will use these new hyperparameters, leading to different classifiers having different hyperparameters. While OvR is retrained from scratch each time a new class enters the system, the other classifiers are updated as discussed in their respective chapters in Section \ref{sec:TheParadigms}. When showing the prediction results at each iteration, only the examples for classes that the model has already seen are used.

\section{Results and Discussion}
In this section the results obtained from the experiments are shown and discussed. The F1-scores for the clusters datasets is always almost identical to the accuracy, since the classes for these datasets are equally distributed, therefore these will not be shown. Time for the clusters datasets will only be shown for one variant as these ended up being very similar. With regards to experiments 1b and 1c, the results for M=0 is the same as that of 1a, hence they are not shown again. The full tables, however, are given in the Appendix \ref{App:FullResultsTables}.

  \begin{table*}
	\centering
	\caption{Experiment 1a - Retraining from scratch}
	\resizebox{\textwidth}{!}{%
		\begin{tabular}{@{}lllllllllll@{}}
			\toprule
			& & & \makecell{OvR\\LR} & \makecell{SCL\\LR} & \makecell{OvP\\LR} & \makecell{OvPSC\\LR} & \makecell{OvR\\SVM} & \makecell{SCL\\SVM} & \makecell{OvP\\SVM} & \makecell{OvPSC\\SVM} \\ \midrule
			\multirow{9}{*}{SpatialVOC2K} & \multirow{3}{*}{M=0} & accuracy & 0.542 & 0.483 & 0.538 & 0.540 & 0.541 & 0.540 & 0.544 & 0.552 \\
			& & f1 & 0.337 & 0.343 & 0.289 & 0.312 & 0.340 & 0.349 & 0.302 & 0.310 \\
			& & time & 0.268 & 1.422 & 0.281 & 0.614 & 1.706 & 10.258 & 0.935 & 1.369 \\ \cmidrule(l){2-11}
			& \multirow{3}{*}{M=1} & accuracy & 0.549 & 0.533 & 0.547 & 0.549 & 0.561 & 0.528 & 0.553 & 0.559 \\
			& & f1 & 0.347 & 0.371 & 0.303 & 0.316 & 0.360 & 0.360 & 0.306 & 0.316 \\
			& & time & 0.429 & 2.332 & 0.435 & 0.980 & 4.233 & 25.382 & 2.195 & 2.820 \\ \cmidrule(l){2-11}
			& \multirow{3}{*}{M=2} & accuracy & 0.548 & 0.542 & 0.554 & 0.555 & 0.567 & 0.526 & 0.564 & 0.567 \\
			& & f1 & 0.351 & 0.367 & 0.316 & 0.327 & 0.365 & 0.363 & 0.327 & 0.317 \\
			& & time & 0.545 & 2.973 & 0.544 & 1.227 & 6.506 & 40.824 & 3.338 & 4.100 \\ \bottomrule
			
			\multirow{6}{*}{clusters\_veryFar} & \multirow{2}{*}{M=0} & accuracy & 1.000 & 0.660 & 0.783 & 0.867 & 1.000 & 1.000 & 1.000 & 1.000 \\
			& & time & 0.024 & 0.147 & 0.035 & 0.104 & 0.028 & 0.163 & 0.035 & 0.119 \\ \cmidrule(l){2-11}
			& \multirow{2}{*}{M=1} & accuracy & 1.000 & 0.660 & 0.783 & 0.761 & 1.000 & 1.000 & 1.000 & 1.000 \\
			& & time & 0.028 & 0.206 & 0.038 & 0.148 & 0.039 & 0.233 & 0.038 & 0.172 \\ \cmidrule(l){2-11}
			& \multirow{2}{*}{M=2} & accuracy & 1.000 & 0.660 & 0.783 & 0.673 & 1.000 & 1.000 & 1.000 & 1.000 \\
			& & time & 0.029 & 0.244 & 0.039 & 0.177 & 0.050 & 0.279 & 0.039 & 0.208 \\ \bottomrule
			
			\multirow{3}{*}{clusters\_far} & \multirow{1}{*}{M=0} & accuracy & 0.978 & 0.745 & 0.918 & 0.920 & 0.991 & 0.999 & 0.996 & 0.996 \\ \cmidrule(l){2-11}
			& \multirow{1}{*}{M=1} & accuracy & 0.983 & 0.712 & 0.925 & 0.925 & 0.988 & 0.999 & 1.000 & 1.000 \\ \cmidrule(l){2-11}
			& \multirow{1}{*}{M=2} & accuracy & 0.984 & 0.702 & 0.928 & 0.930 & 0.980 & 1.000 & 1.000 & 1.000 \\ \bottomrule
		\end{tabular}%
	}
	\label{Table:1a}
\end{table*}

\begin{table*}
	\centering
	\caption{Experiment 1b - Fine-tune with new set only}
	\resizebox{\textwidth}{!}{%
		\begin{tabular}{@{}lllllllllll@{}}
			\toprule
			& & & \makecell{OvR\\LR} & \makecell{SCL\\LR} & \makecell{OvP\\LR} & \makecell{OvPSC\\LR} & \makecell{OvR\\SVM} & \makecell{SCL\\SVM} & \makecell{OvP\\SVM} & \makecell{OvPSC\\SVM} \\ \midrule
			\multirow{6}{*}{SpatialVOC2K} & \multirow{3}{*}{M=1} & accuracy & 0.532 & 0.470 & 0.531 & 0.527 & 0.530 & 0.518 & 0.528 & 0.540 \\
			& & f1 & 0.292 & 0.304 & 0.282 & 0.288 & 0.315 & 0.303 & 0.286 & 0.290 \\
			& & time & 0.162 & 1.051 & 0.192 & 0.130 & 0.658 & 5.706 & 0.395 & 0.291 \\ \cmidrule(l){2-11}
			& \multirow{3}{*}{M=2} & accuracy & 0.528 & 0.456 & 0.524 & 0.520 & 0.520 & 0.515 & 0.521 & 0.531 \\
			& & f1 & 0.286 & 0.287 & 0.274 & 0.287 & 0.302 & 0.291 & 0.284 & 0.276 \\
			& & time & 0.111 & 0.740 & 0.136 & 0.091 & 0.327 & 2.740 & 0.219 & 0.154 \\ \bottomrule
			
			\multirow{4}{*}{clusters\_far} & \multirow{2}{*}{M=1} & accuracy & 0.980 & 0.669 & 0.918 & 0.918 & 0.977 & 0.984 & 1.000 & 0.999 \\
			& & time & 0.020 & 0.076 & 0.032 & 0.017 & 0.020 & 0.070 & 0.032 & 0.016 \\ \cmidrule(l){2-11}
			& \multirow{2}{*}{M=2} & accuracy & 0.977 & 0.682 & 0.918 & 0.907 & 0.955 & 0.994 & 0.993 & 0.980 \\
			& & time & 0.018 & 0.074 & 0.032 & 0.016 & 0.019 & 0.066 & 0.031 & 0.016 \\ \bottomrule
			
		\end{tabular}%
	}
	\label{Table:1b}
\end{table*}

\begin{table*}
	\centering
	\caption{Experiment 1c - Fine-tune with entire dataset} 
	\resizebox{\textwidth}{!}{%
		\begin{tabular}{@{}lllllllllll@{}}
			\toprule
			& & & \makecell{OvR\\LR} & \makecell{SCL\\LR} & \makecell{OvP\\LR} & \makecell{OvPSC\\LR} & \makecell{OvR\\SVM} & \makecell{SCL\\SVM} & \makecell{OvP\\SVM} & \makecell{OvPSC\\SVM} \\ \midrule
			\multirow{6}{*}{SpatialVOC2K} & \multirow{3}{*}{M=1} & accuracy & 0.549 & 0.486 & 0.547 & 0.548 & 0.561 & 0.556 & 0.553 & 0.559 \\
			& & f1 & 0.347 & 0.365 & 0.303 & 0.313 & 0.360 & 0.358 & 0.306 & 0.312 \\
			& & time & 0.423 & 2.685 & 0.432 & 0.341 & 4.379 & 37.415 & 2.181 & 1.698 \\ \cmidrule(l){2-11}
			& \multirow{3}{*}{M=2} & accuracy & 0.548 & 0.485 & 0.554 & 0.555 & 0.567 & 0.561 & 0.564 & 0.567 \\
			& & f1 & 0.351 & 0.355 & 0.316 & 0.326 & 0.365 & 0.365 & 0.327 & 0.314 \\
			& & time & 0.541 & 3.263 & 0.542 & 0.431 & 6.702 & 58.650 & 3.308 & 2.557 \\ \bottomrule
			
			\multirow{4}{*}{clusters\_far} & \multirow{2}{*}{M=1} & accuracy & 0.983 & 0.712 & 0.925 & 0.925 & 0.988 & 0.997 & 1.000 & 0.987 \\
			& & time & 0.025 & 0.088 & 0.036 & 0.019 & 0.032 & 0.085 & 0.035 & 0.019 \\ \cmidrule(l){2-11}
			& \multirow{2}{*}{M=2} & accuracy & 0.984 & 0.702 & 0.928 & 0.928 & 0.980 & 0.997 & 1.000 & 0.983 \\
			& & time & 0.027 & 0.093 & 0.038 & 0.020 & 0.038 & 0.092 & 0.038 & 0.020 \\ \bottomrule
		\end{tabular}%
	}
\label{Table:1c}
\end{table*}

\begin{table*}
	\centering
	\caption{Experiment 2 - Retrain with new class - SpatialVOC2K - Last Class}
	\resizebox{\textwidth}{!}{%
		\begin{tabular}{@{}llllllllll@{}}
			\toprule
			& & \makecell{OvR\\LR} & \makecell{SCL\\LR} & \makecell{OvP\\LR} & \makecell{OvPSC\\LR} & \makecell{OvR\\SVM} & \makecell{SCL\\SVM} & \makecell{OvP\\SVM} & \makecell{OvPSC\\SVM} \\ \midrule
			\multirow{3}{*}{N17=par\_dela} & accuracy & 0.547 & 0.537 & 0.553 & 0.554 & 0.573 & 0.442 & 0.564 & 0.563 \\
			& f1 & 0.351 & 0.341 & 0.310 & 0.308 & 0.348 & 0.191 & 0.303 & 0.299 \\
			& time & 0.640 & 0.090 & 0.011 & 0.022 & 5.695 & 0.909 & 0.010 & 0.026 \\ \bottomrule
		\end{tabular}%
	}
\label{Table:2-SpatRel}
\end{table*}

\begin{table*}
	\centering
	\caption{Experiment 2 - Retrain with new class - clusters\_far - Last class}
	\resizebox{\textwidth}{!}{%
		\begin{tabular}{@{}llllllllll@{}}
			\toprule
			& & \makecell{OvR\\LR} & \makecell{SCL\\LR} & \makecell{OvP\\LR} & \makecell{OvPSC\\LR} & \makecell{OvR\\SVM} & \makecell{SCL\\SVM} & \makecell{OvP\\SVM} & \makecell{OvPSC\\SVM} \\ \midrule
			\multirow{3}{*}{N9=2} & accuracy & 0.981 & 0.689 & 0.914 & 0.921 & 0.997 & 0.588 & 0.906 & 0.816 \\
			& f1 & 0.982 & 0.691 & 0.910 & 0.918 & 0.997 & 0.488 & 0.881 & 0.746 \\
			& time & 0.029 & 0.048 & 0.002 & 0.038 & 0.038 & 0.057 & 0.003 & 0.047 \\ \bottomrule
		\end{tabular}%
	}
\label{Table:2-circles}
\end{table*}

  \subsection{Experiment 1}
\subsubsection{Experiment 1a - Retraining from scratch}
Table \ref{Table:1a} shows that Logistic Regression, in terms of classification performance, performs worse on the \textit{clusters} datasets when compared to SVM, as this dataset is not-linearly separable. Meanwhile, it performs only slightly worse on SpatialVOC2K, indicating that the features of this dataset make it more linearly separable. Almost every classifier gets better classification performance as the dataset size increases, that is, as M increases. Subsequently, it also takes more time to train. As expected, LR also takes less time to train than the corresponding SVM models, when the paradigm is kept constant.

The three proposed paradigms are comparable to the baseline in classification performance, in some cases being slightly better. That said, OvR is still usually better, albeit with a small difference ranging from 1\%-5\%. SCL-LR when used on the \textit{clusters} dataset, however, performs poorly with regards to classification performance. This can be seen in the \textit{clusters\_veryFar} dataset results. With regards to the F1 score on SpatialVOC2K, the most comparable to OvR is the SCL paradigm, while OvP and OvPSC perform around 4-5\% worse. Meanwhile, OvP and OvPSC accuracy is comparable to that of OvR, with less than 1\% difference.

With regards to time, SCL always takes the longest when given the entire dataset as a whole, due to the retraining this paradigm performs. Focusing on the SVM variants, OvP is always the one which takes the least time, taking half the time of OvR. This takes less time than OvPSC, even though the latter uses less training examples, due to OvP not needing to check for similar classes, which is an indication that this checking contributes a significant overhead. OvR is faster than OvPSC on the \textit{clusters} datasets, further solidifying the fact that this overhead is significant. However, on SpatialVOC2K OvPSC is faster than OvR, which indicates that OvPSC scales better on bigger datasets, as the overhead becomes less significant.

\subsubsection{Experiment 1b - Fine-tune with new set}
In this experiment, SCL-LR also performs poorly on the \textit{clusters} datasets (Table \ref{Table:1b}). In terms of the Logistic Regression models, OvR is the best performing paradigm on the \textit{clusters} Datasets. Since there is less data in each M-set, training time for all models takes less each time, indicating that the models do not take longer to converge if different data is used. Furthermore, OvPSC takes less time than OvR in this case, because it does not need to search for similar classes again. When considering the SVM models, in most cases, at least one of the models built with one of the proposed paradigms either outperforms, is equal to or comes very close to OvR with regards to classification performance on SpatialVOC2K, while the time taken to train is much less. When all M-sets enter the system, almost no classifier performed better than that in experiment 1a, indicating that for these datasets, fine-tuning with the new set only might not be as effective in retaining classification performance. However, if time has a higher priority, one will have to consider the trade-off, especially on much larger datasets.

\subsubsection{Experiment 1c - Fine-tune with entire dataset}
After conducting this experiment, with regards to time, OvR and OvP remain mostly unaffected, while OvPSC takes less time. On the other hand, the SCL-SVM model, takes much longer on SpatialVOC2K as M increases but ends up improving its classification performance over experiment 1a. Classification performance for the other models is mostly unaffected. No significant difference was observed on the clusters datasets.

  \subsection{Experiment 2}
In this section, the results after training for the final class are shown in Tables \ref{Table:2-SpatRel} and \ref{Table:2-circles} for SpatialVOC2K and the \textit{clusters\_far} dataset respectively.
It can be observed that, for SpatialVOC2K especially, timing to add a new class decreases remarkably when using one of the proposed paradigms, since they do not retrain from scratch.
Classification performance for the Logistic Regression models are retained when compared to previous experiments.
While OvR retains its classification performance when using SVM on the \textit{clusters} dataset due to retraining from scratch, the other methods perform worse when using different hyperparameters for different classifiers on this dataset.
When comparing the SVM variants on SpatialVOC2K, while SCL performs worse when changing hyperparameters between classifiers, classification performance for OvP and OvPSC is mostly retained, with about 1\% less accuracy and 5\% less F1-score.

\section{Conclusion}
This work proposed three paradigms to train models which can be used with any model comprised of a series of binary classifiers. These are called \textit{Similar Classes Learning} (SCL), \textit{One vs Previous} (OvP) and \textit{One vs Previous and Similar Classes} (OvPSC). The main purpose of these paradigms is to be more efficient than existing paradigms while retaining classification performance. Detailed descriptions of these paradigms were given through their respective pseudo-code algorithms. Two datasets were used: SpatialVOC2K, which is a real-world dataset used to show the paradigms at work in a practical scenario and a number of synthetic datasets to see how the paradigms vary across these. A comparative analysis was made which compared these datasets to a standard paradigm.

The conducted experiments evaluated both the classification as well as computational performance of the paradigms. Results show that the new paradigms are faster at retraining than the baselines as new classes come in, with OvP being faster at training overall, while OvPSC being faster as datasets scale up. This means that OvP and OvPSC can be used as a replacement of OvR for faster training and prediction, if the classification performance is not affected. In most cases, the classification performance was similar to the baseline, being in the range of 1\%-5\%. Therefore, the learning paradigm may be a considerable variable when training a machine learning model, depending on the dataset and application of the model.

\subsection{Future Work}
This section outlines improvements which can make the paradigms more widely applicable. 

Firstly, as was discussed, the order in which classes enter the system determines the shape of the decision boundaries of the models built. While ordering in most frequent first was shown to give good results, there may be other class combinations which give better results. Therefore, a study which determines the optimal order could be carried out. 

Second, the implementation of the proposed paradigms was carried out without many resources dedicated to optimization. Hence, more optimization could further decrease the time taken to train and predict by these paradigms.

Lastly, adopting these paradigms to Neural Networks would make them more applicable to the current state of machine learning, while addressing the problem of long training times of complex Neural Networks.

\bibliographystyle{unsrt}  
\bibliography{ms}

\onecolumn
\appendix
\section{SpatialVOC2K Additional Details and Visualisations}

\textbf{Full list of chosen geometric features:} \sloppy{\textit{AreaObj1\_Norm\_wt\_Union, AreaObj2\_Norm\_wt\_Union, AreaOverlap\_Norm\_wt\_Union, AspectRatioObj\_1, AspectRatioObj\_2, DistBtCentr\_Norm\_wt\_UnionBB, DistanceSizeRatio, EuclDist\_Norm\_wt\_UnionBB, InvFeatXmaxXmin, InvFeatXminXmin, InvFeatYmaxYmin, InvFeatYminYmin, objAreaRatioMaxMin, objAreaRatioTrajLand, relativePosition\_one\_hot\_0, relativePosition\_one\_hot\_1, relativePosition\_one\_hot\_2, relativePosition\_one\_hot\_3, vecTrajLand\_Norm\_wt\_UnionBB\_x, vecTrajLand\_Norm\_wt\_UnionBB\_y, depth\_human\_0, depth\_human\_1, depth\_human\_2}} \label{App:SpatRelFeatures}

\begin{figure}[H]
	\centering
	\includegraphics[width=1\textwidth]{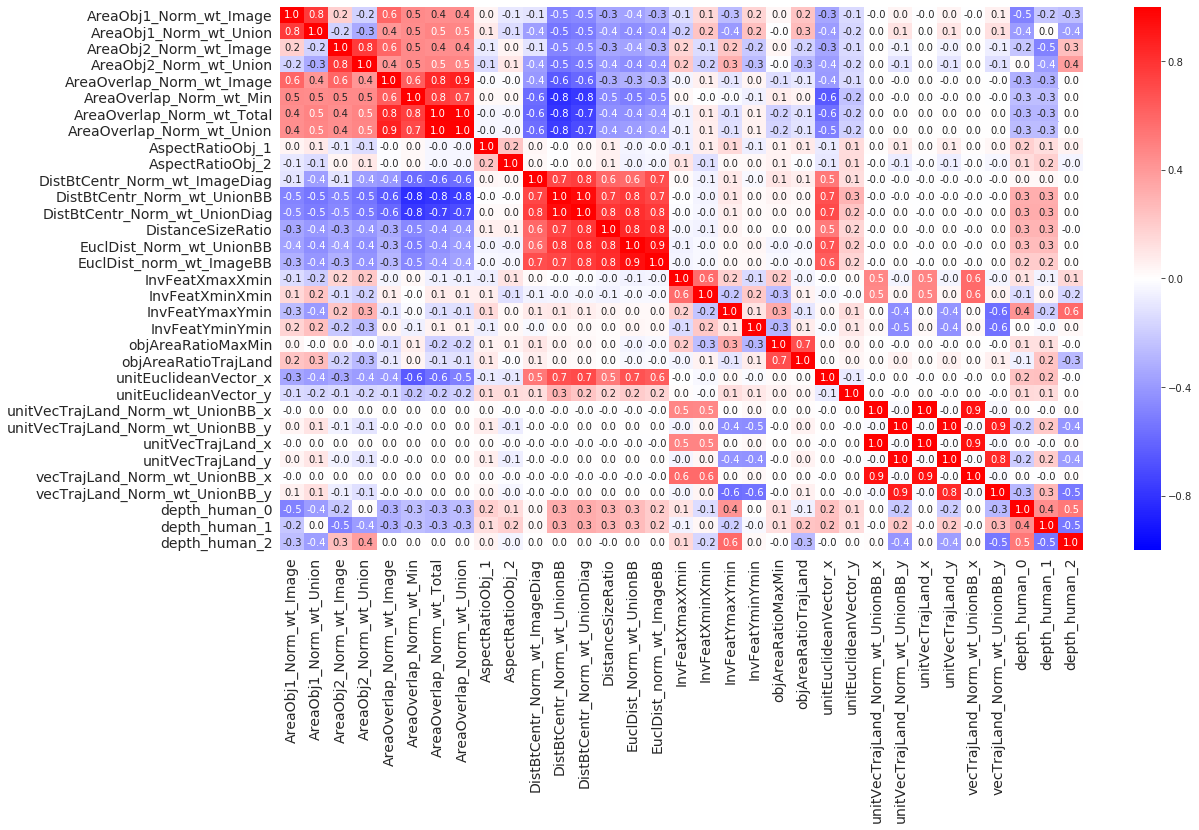}
	\caption{Correlation Matrix of Geometric Features}
\end{figure}

\begin{figure}[H]
	\centering
	\includegraphics[width=0.8\textwidth]{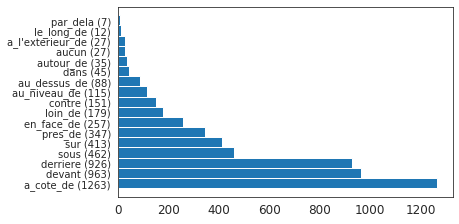}
	\caption{The frequencies of the SpatialVOC2K targets}
	\label{fig:SpatRel_distribution}
\end{figure}

\begin{figure}[H]
	\centering
	\includegraphics[width=\textwidth]{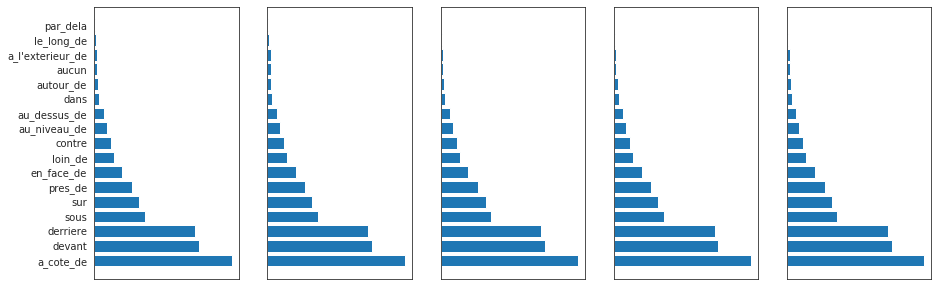}
	\caption{Distribution of the 5 stratified folds}
\end{figure}

\section{Synthetic Clusters Additional Visualisations}
\begin{figure}[H]
	\centering
	\includegraphics[width=.8\textwidth]{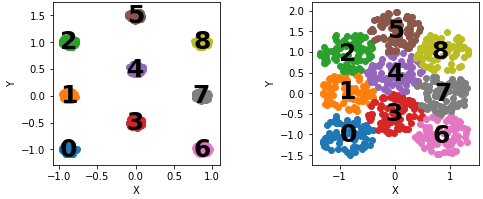}
	\caption{The synthetic clusters datasets with radius=0.1 (left) and radius=0.5 (right). The numbers represent the class labels.}
\end{figure} \label{App:Clusters}

\begin{figure}[H]
	\centering
	\includegraphics[width=.6\textwidth]{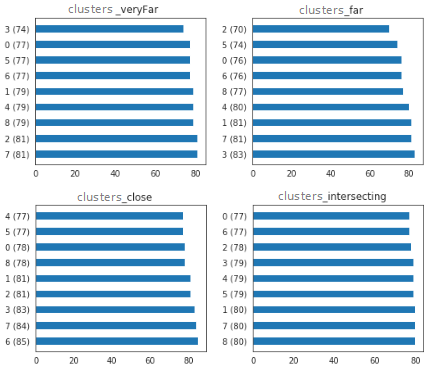}
	\caption{Frequency bar plot of each of the \textit{clusters} datasets}
\end{figure}

\section{Experiments Full Results Tables}
\label{App:FullResultsTables}
\begin{table}[H]
	\centering
	\caption{Experiment 1a - Retrain from scratch}
	\resizebox{0.64\textwidth}{!}{%
		\begin{tabular}{@{}lllllllllll@{}}
			\toprule
			& & & \makecell{OvR\\LR} & \makecell{SCL\\LR} & \makecell{OvP\\LR} & \makecell{OvPSC\\LR} & \makecell{OvR\\SVM} & \makecell{SCL\\SVM} & \makecell{OvP\\SVM} & \makecell{OvPSC\\SVM} \\ \midrule
			\multirow{9}{*}{SpatialRelations} & \multirow{3}{*}{M=0} & accuracy & 0.542 & 0.483 & 0.538 & 0.540 & 0.541 & 0.540 & 0.544 & 0.552 \\
			& & f1 & 0.337 & 0.343 & 0.289 & 0.312 & 0.340 & 0.349 & 0.302 & 0.310 \\
			& & time & 0.268 & 1.422 & 0.281 & 0.614 & 1.706 & 10.258 & 0.935 & 1.369 \\ \cmidrule(l){2-11}
			& \multirow{3}{*}{M=1} & accuracy & 0.549 & 0.533 & 0.547 & 0.549 & 0.561 & 0.528 & 0.553 & 0.559 \\
			& & f1 & 0.347 & 0.371 & 0.303 & 0.316 & 0.360 & 0.360 & 0.306 & 0.316 \\
			& & time & 0.429 & 2.332 & 0.435 & 0.980 & 4.233 & 25.382 & 2.195 & 2.820 \\ \cmidrule(l){2-11}
			& \multirow{3}{*}{M=2} & accuracy & 0.548 & 0.542 & 0.554 & 0.555 & 0.567 & 0.526 & 0.564 & 0.567 \\
			& & f1 & 0.351 & 0.367 & 0.316 & 0.327 & 0.365 & 0.363 & 0.327 & 0.317 \\
			& & time & 0.545 & 2.973 & 0.544 & 1.227 & 6.506 & 40.824 & 3.338 & 4.100 \\ \bottomrule
			\multirow{9}{*}{clusters\_veryFar} & \multirow{3}{*}{M=0} & accuracy & 1.000 & 0.660 & 0.783 & 0.867 & 1.000 & 1.000 & 1.000 & 1.000 \\
			& & f1 & 1.000 & 0.598 & 0.701 & 0.831 & 1.000 & 1.000 & 1.000 & 1.000 \\
			& & time & 0.024 & 0.147 & 0.035 & 0.104 & 0.028 & 0.163 & 0.035 & 0.119 \\ \cmidrule(l){2-11}
			& \multirow{3}{*}{M=1} & accuracy & 1.000 & 0.660 & 0.783 & 0.761 & 1.000 & 1.000 & 1.000 & 1.000 \\
			& & f1 & 1.000 & 0.598 & 0.700 & 0.679 & 1.000 & 1.000 & 1.000 & 1.000 \\
			& & time & 0.028 & 0.206 & 0.038 & 0.148 & 0.039 & 0.233 & 0.038 & 0.172 \\ \cmidrule(l){2-11}
			& \multirow{3}{*}{M=2} & accuracy & 1.000 & 0.660 & 0.783 & 0.673 & 1.000 & 1.000 & 1.000 & 1.000 \\
			& & f1 & 1.000 & 0.598 & 0.700 & 0.559 & 1.000 & 1.000 & 1.000 & 1.000 \\
			& & time & 0.029 & 0.244 & 0.039 & 0.177 & 0.050 & 0.279 & 0.039 & 0.208 \\ \bottomrule
			\multirow{9}{*}{clusters\_far} & \multirow{3}{*}{M=0} & accuracy & 0.978 & 0.745 & 0.918 & 0.920 & 0.991 & 0.999 & 0.996 & 0.996 \\
			& & f1 & 0.978 & 0.719 & 0.916 & 0.917 & 0.991 & 0.999 & 0.996 & 0.996 \\
			& & time & 0.024 & 0.149 & 0.036 & 0.102 & 0.027 & 0.166 & 0.035 & 0.116 \\ \cmidrule(l){2-11}
			& \multirow{3}{*}{M=1} & accuracy & 0.983 & 0.712 & 0.925 & 0.925 & 0.988 & 0.999 & 1.000 & 1.000 \\
			& & f1 & 0.983 & 0.714 & 0.923 & 0.924 & 0.988 & 0.999 & 1.000 & 1.000 \\
			& & time & 0.027 & 0.208 & 0.039 & 0.144 & 0.034 & 0.241 & 0.038 & 0.171 \\ \cmidrule(l){2-11}
			& \multirow{3}{*}{M=2} & accuracy & 0.984 & 0.702 & 0.928 & 0.930 & 0.980 & 1.000 & 1.000 & 1.000 \\
			& & f1 & 0.984 & 0.717 & 0.926 & 0.928 & 0.980 & 1.000 & 1.000 & 1.000 \\
			& & time & 0.030 & 0.244 & 0.040 & 0.172 & 0.041 & 0.284 & 0.039 & 0.207 \\ \bottomrule
			\multirow{9}{*}{clusters\_close} & \multirow{3}{*}{M=0} & accuracy & 0.898 & 0.573 & 0.801 & 0.814 & 0.982 & 0.978 & 0.975 & 0.975 \\
			& & f1 & 0.896 & 0.535 & 0.751 & 0.775 & 0.982 & 0.978 & 0.975 & 0.975 \\
			& & time & 0.024 & 0.145 & 0.035 & 0.105 & 0.027 & 0.175 & 0.036 & 0.121 \\ \cmidrule(l){2-11}
			& \multirow{3}{*}{M=1} & accuracy & 0.909 & 0.584 & 0.804 & 0.812 & 0.986 & 0.988 & 0.979 & 0.979 \\
			& & f1 & 0.908 & 0.550 & 0.756 & 0.772 & 0.986 & 0.987 & 0.979 & 0.979 \\
			& & time & 0.028 & 0.207 & 0.038 & 0.151 & 0.032 & 0.249 & 0.039 & 0.179 \\ \cmidrule(l){2-11}
			& \multirow{3}{*}{M=2} & accuracy & 0.908 & 0.573 & 0.803 & 0.809 & 0.986 & 0.989 & 0.988 & 0.988 \\
			& & f1 & 0.906 & 0.535 & 0.754 & 0.772 & 0.987 & 0.989 & 0.988 & 0.988 \\
			& & time & 0.030 & 0.246 & 0.040 & 0.181 & 0.037 & 0.302 & 0.040 & 0.216 \\ \bottomrule
			\multirow{9}{*}{clusters\_intersecting} & \multirow{3}{*}{M=0} & accuracy & 0.900 & 0.710 & 0.709 & 0.719 & 0.958 & 0.955 & 0.938 & 0.937 \\
			& & f1 & 0.897 & 0.689 & 0.690 & 0.698 & 0.958 & 0.955 & 0.938 & 0.936 \\
			& & time & 0.024 & 0.154 & 0.035 & 0.104 & 0.027 & 0.177 & 0.036 & 0.122 \\ \cmidrule(l){2-11}
			& \multirow{3}{*}{M=1} & accuracy & 0.904 & 0.694 & 0.715 & 0.722 & 0.967 & 0.959 & 0.948 & 0.948 \\
			& & f1 & 0.901 & 0.683 & 0.697 & 0.700 & 0.968 & 0.959 & 0.948 & 0.948 \\
			& & time & 0.028 & 0.212 & 0.038 & 0.149 & 0.032 & 0.254 & 0.039 & 0.180 \\ \cmidrule(l){2-11}
			& \multirow{3}{*}{M=2} & accuracy & 0.894 & 0.674 & 0.723 & 0.732 & 0.968 & 0.959 & 0.955 & 0.953 \\
			& & f1 & 0.892 & 0.663 & 0.700 & 0.703 & 0.968 & 0.959 & 0.955 & 0.953 \\
			& & time & 0.029 & 0.251 & 0.039 & 0.179 & 0.037 & 0.302 & 0.042 & 0.215 \\ \bottomrule
		\end{tabular}%
	}
\end{table}
\begin{table}[H]
	\centering
	\caption{Experiment 1b - Fine-tune with new set only} 
	\resizebox{0.68\textwidth}{!}{%
		\begin{tabular}{@{}lllllllllll@{}}
			\toprule
			& & & \makecell{OvR\\LR} & \makecell{SCL\\LR} & \makecell{OvP\\LR} & \makecell{OvPSC\\LR} & \makecell{OvR\\SVM} & \makecell{SCL\\SVM} & \makecell{OvP\\SVM} & \makecell{OvPSC\\SVM} \\ \midrule
			\multirow{9}{*}{SpatialRelations} & \multirow{3}{*}{M=0} & accuracy & 0.542 & 0.483 & 0.538 & 0.540 & 0.541 & 0.540 & 0.544 & 0.552 \\
			& & f1 & 0.337 & 0.343 & 0.289 & 0.312 & 0.340 & 0.349 & 0.302 & 0.310 \\
			& & time & 0.266 & 1.441 & 0.289 & 0.624 & 1.713 & 10.374 & 0.936 & 1.388 \\ \cmidrule(l){2-11}
			& \multirow{3}{*}{M=1} & accuracy & 0.532 & 0.470 & 0.531 & 0.527 & 0.530 & 0.518 & 0.528 & 0.540 \\
			& & f1 & 0.292 & 0.304 & 0.282 & 0.288 & 0.315 & 0.303 & 0.286 & 0.290 \\
			& & time & 0.162 & 1.051 & 0.192 & 0.130 & 0.658 & 5.706 & 0.395 & 0.291 \\ \cmidrule(l){2-11}
			& \multirow{3}{*}{M=2} & accuracy & 0.528 & 0.456 & 0.524 & 0.520 & 0.520 & 0.515 & 0.521 & 0.531 \\
			& & f1 & 0.286 & 0.287 & 0.274 & 0.287 & 0.302 & 0.291 & 0.284 & 0.276 \\
			& & time & 0.111 & 0.740 & 0.136 & 0.091 & 0.327 & 2.740 & 0.219 & 0.154 \\ \bottomrule
			\multirow{9}{*}{clusters\_veryFar} & \multirow{3}{*}{M=0} & accuracy & 1.000 & 0.660 & 0.783 & 0.867 & 1.000 & 1.000 & 1.000 & 1.000 \\
			& & f1 & 1.000 & 0.598 & 0.701 & 0.831 & 1.000 & 1.000 & 1.000 & 1.000 \\
			& & time & 0.024 & 0.147 & 0.035 & 0.105 & 0.028 & 0.165 & 0.035 & 0.124 \\ \cmidrule(l){2-11}
			& \multirow{3}{*}{M=1} & accuracy & 1.000 & 0.429 & 0.766 & 0.776 & 1.000 & 1.000 & 1.000 & 0.994 \\
			& & f1 & 1.000 & 0.396 & 0.684 & 0.717 & 1.000 & 1.000 & 1.000 & 0.994 \\
			& & time & 0.020 & 0.066 & 0.032 & 0.016 & 0.022 & 0.063 & 0.032 & 0.016 \\ \cmidrule(l){2-11}
			& \multirow{3}{*}{M=2} & accuracy & 1.000 & 0.409 & 0.699 & 0.776 & 1.000 & 1.000 & 1.000 & 0.983 \\
			& & f1 & 1.000 & 0.370 & 0.601 & 0.717 & 1.000 & 1.000 & 1.000 & 0.983 \\
			& & time & 0.021 & 0.064 & 0.032 & 0.016 & 0.019 & 0.059 & 0.031 & 0.016 \\ \bottomrule
			\multirow{9}{*}{clusters\_far} & \multirow{3}{*}{M=0} & accuracy & 0.978 & 0.745 & 0.918 & 0.920 & 0.991 & 0.999 & 0.996 & 0.996 \\
			& & f1 & 0.978 & 0.719 & 0.916 & 0.917 & 0.991 & 0.999 & 0.996 & 0.996 \\
			& & time & 0.025 & 0.151 & 0.036 & 0.102 & 0.027 & 0.167 & 0.036 & 0.118 \\ \cmidrule(l){2-11}
			& \multirow{3}{*}{M=1} & accuracy & 0.980 & 0.669 & 0.918 & 0.918 & 0.977 & 0.984 & 1.000 & 0.999 \\
			& & f1 & 0.980 & 0.699 & 0.916 & 0.916 & 0.977 & 0.985 & 1.000 & 0.999 \\
			& & time & 0.020 & 0.076 & 0.032 & 0.017 & 0.020 & 0.070 & 0.032 & 0.016 \\ \cmidrule(l){2-11}
			& \multirow{3}{*}{M=2} & accuracy & 0.977 & 0.682 & 0.918 & 0.907 & 0.955 & 0.994 & 0.993 & 0.980 \\
			& & f1 & 0.978 & 0.710 & 0.916 & 0.905 & 0.954 & 0.994 & 0.993 & 0.980 \\
			& & time & 0.018 & 0.074 & 0.032 & 0.016 & 0.019 & 0.066 & 0.031 & 0.016 \\ \bottomrule
			\multirow{9}{*}{clusters\_close} & \multirow{3}{*}{M=0} & accuracy & 0.898 & 0.573 & 0.801 & 0.814 & 0.982 & 0.978 & 0.975 & 0.975 \\
			& & f1 & 0.896 & 0.535 & 0.751 & 0.775 & 0.982 & 0.978 & 0.975 & 0.975 \\
			& & time & 0.024 & 0.148 & 0.037 & 0.106 & 0.027 & 0.178 & 0.035 & 0.124 \\ \cmidrule(l){2-11}
			& \multirow{3}{*}{M=1} & accuracy & 0.909 & 0.543 & 0.801 & 0.804 & 0.978 & 0.971 & 0.963 & 0.950 \\
			& & f1 & 0.908 & 0.523 & 0.752 & 0.771 & 0.977 & 0.971 & 0.962 & 0.949 \\
			& & time & 0.020 & 0.070 & 0.033 & 0.017 & 0.020 & 0.076 & 0.032 & 0.016 \\ \cmidrule(l){2-11}
			& \multirow{3}{*}{M=2} & accuracy & 0.867 & 0.561 & 0.772 & 0.751 & 0.927 & 0.920 & 0.926 & 0.914 \\
			& & f1 & 0.866 & 0.523 & 0.720 & 0.714 & 0.923 & 0.916 & 0.925 & 0.914 \\
			& & time & 0.019 & 0.069 & 0.032 & 0.016 & 0.019 & 0.073 & 0.032 & 0.015 \\ \bottomrule
			\multirow{9}{*}{clusters\_intersecting} & \multirow{3}{*}{M=0} & accuracy & 0.900 & 0.710 & 0.709 & 0.719 & 0.958 & 0.955 & 0.938 & 0.937 \\
			& & f1 & 0.897 & 0.689 & 0.690 & 0.698 & 0.958 & 0.955 & 0.938 & 0.936 \\
			& & time & 0.024 & 0.158 & 0.036 & 0.106 & 0.026 & 0.179 & 0.036 & 0.124 \\ \cmidrule(l){2-11}
			& \multirow{3}{*}{M=1} & accuracy & 0.896 & 0.632 & 0.684 & 0.705 & 0.928 & 0.921 & 0.901 & 0.906 \\
			& & f1 & 0.893 & 0.627 & 0.660 & 0.685 & 0.928 & 0.921 & 0.901 & 0.906 \\
			& & time & 0.020 & 0.079 & 0.033 & 0.016 & 0.020 & 0.078 & 0.032 & 0.017 \\ \cmidrule(l){2-11}
			& \multirow{3}{*}{M=2} & accuracy & 0.880 & 0.606 & 0.685 & 0.724 & 0.935 & 0.903 & 0.907 & 0.904 \\
			& & f1 & 0.877 & 0.555 & 0.642 & 0.686 & 0.935 & 0.904 & 0.907 & 0.905 \\
			& & time & 0.019 & 0.077 & 0.032 & 0.016 & 0.018 & 0.081 & 0.031 & 0.016 \\ \bottomrule
		\end{tabular}%
	}
\end{table}
\begin{table}[H]
	\centering
	\caption{Experiment 1c - Fine-tune with entire dataset} 
	\resizebox{0.68\textwidth}{!}{%
		\begin{tabular}{@{}lllllllllll@{}}
			\toprule
			& & & \makecell{OvR\\LR} & \makecell{SCL\\LR} & \makecell{OvP\\LR} & \makecell{OvPSC\\LR} & \makecell{OvR\\SVM} & \makecell{SCL\\SVM} & \makecell{OvP\\SVM} & \makecell{OvPSC\\SVM} \\ \midrule
			\multirow{9}{*}{SpatialRelations} & \multirow{3}{*}{M=0} & accuracy & 0.542 & 0.483 & 0.538 & 0.540 & 0.541 & 0.540 & 0.544 & 0.552 \\
			& & f1 & 0.337 & 0.343 & 0.289 & 0.312 & 0.340 & 0.349 & 0.302 & 0.310 \\
			& & time & 0.270 & 1.523 & 0.284 & 0.661 & 1.794 & 10.298 & 0.931 & 1.372 \\ \cmidrule(l){2-11}
			& \multirow{3}{*}{M=1} & accuracy & 0.549 & 0.486 & 0.547 & 0.548 & 0.561 & 0.556 & 0.553 & 0.559 \\
			& & f1 & 0.347 & 0.365 & 0.303 & 0.313 & 0.360 & 0.358 & 0.306 & 0.312 \\
			& & time & 0.423 & 2.685 & 0.432 & 0.341 & 4.379 & 37.415 & 2.181 & 1.698 \\ \cmidrule(l){2-11}
			& \multirow{3}{*}{M=2} & accuracy & 0.548 & 0.485 & 0.554 & 0.555 & 0.567 & 0.561 & 0.564 & 0.567 \\
			& & f1 & 0.351 & 0.355 & 0.316 & 0.326 & 0.365 & 0.365 & 0.327 & 0.314 \\
			& & time & 0.541 & 3.263 & 0.542 & 0.431 & 6.702 & 58.650 & 3.308 & 2.557 \\ \bottomrule
			\multirow{9}{*}{clusters\_veryFar} & \multirow{3}{*}{M=0} & accuracy & 1.000 & 0.660 & 0.783 & 0.867 & 1.000 & 1.000 & 1.000 & 1.000 \\
			& & f1 & 1.000 & 0.598 & 0.701 & 0.831 & 1.000 & 1.000 & 1.000 & 1.000 \\
			& & time & 0.024 & 0.146 & 0.034 & 0.102 & 0.028 & 0.164 & 0.036 & 0.118 \\ \cmidrule(l){2-11}
			& \multirow{3}{*}{M=1} & accuracy & 1.000 & 0.596 & 0.783 & 0.879 & 1.000 & 1.000 & 1.000 & 1.000 \\
			& & f1 & 1.000 & 0.531 & 0.700 & 0.838 & 1.000 & 1.000 & 1.000 & 1.000 \\
			& & time & 0.024 & 0.077 & 0.036 & 0.018 & 0.037 & 0.073 & 0.035 & 0.019 \\ \cmidrule(l){2-11}
			& \multirow{3}{*}{M=2} & accuracy & 1.000 & 0.616 & 0.783 & 0.839 & 1.000 & 1.000 & 1.000 & 1.000 \\
			& & f1 & 1.000 & 0.556 & 0.700 & 0.797 & 1.000 & 1.000 & 1.000 & 1.000 \\
			& & time & 0.027 & 0.082 & 0.037 & 0.019 & 0.047 & 0.077 & 0.037 & 0.019 \\ \bottomrule
			\multirow{9}{*}{clusters\_far} & \multirow{3}{*}{M=0} & accuracy & 0.978 & 0.745 & 0.918 & 0.920 & 0.991 & 0.999 & 0.996 & 0.996 \\
			& & f1 & 0.978 & 0.719 & 0.916 & 0.917 & 0.991 & 0.999 & 0.996 & 0.996 \\
			& & time & 0.024 & 0.150 & 0.036 & 0.104 & 0.026 & 0.168 & 0.035 & 0.116 \\ \cmidrule(l){2-11}
			& \multirow{3}{*}{M=1} & accuracy & 0.983 & 0.712 & 0.925 & 0.925 & 0.988 & 0.997 & 1.000 & 0.987 \\
			& & f1 & 0.983 & 0.714 & 0.923 & 0.923 & 0.988 & 0.997 & 1.000 & 0.987 \\
			& & time & 0.025 & 0.088 & 0.036 & 0.019 & 0.032 & 0.085 & 0.035 & 0.019 \\ \cmidrule(l){2-11}
			& \multirow{3}{*}{M=2} & accuracy & 0.984 & 0.702 & 0.928 & 0.928 & 0.980 & 0.997 & 1.000 & 0.983 \\
			& & f1 & 0.984 & 0.717 & 0.926 & 0.926 & 0.980 & 0.997 & 1.000 & 0.982 \\
			& & time & 0.027 & 0.093 & 0.038 & 0.020 & 0.038 & 0.092 & 0.038 & 0.020 \\ \bottomrule
			\multirow{9}{*}{clusters\_close} & \multirow{3}{*}{M=0} & accuracy & 0.898 & 0.573 & 0.801 & 0.814 & 0.982 & 0.978 & 0.975 & 0.975 \\
			& & f1 & 0.896 & 0.535 & 0.751 & 0.775 & 0.982 & 0.978 & 0.975 & 0.975 \\
			& & time & 0.024 & 0.146 & 0.035 & 0.105 & 0.026 & 0.176 & 0.035 & 0.131 \\ \cmidrule(l){2-11}
			& \multirow{3}{*}{M=1} & accuracy & 0.909 & 0.577 & 0.804 & 0.812 & 0.986 & 0.983 & 0.979 & 0.979 \\
			& & f1 & 0.908 & 0.544 & 0.756 & 0.777 & 0.986 & 0.983 & 0.979 & 0.979 \\
			& & time & 0.026 & 0.082 & 0.037 & 0.020 & 0.029 & 0.096 & 0.036 & 0.019 \\ \cmidrule(l){2-11}
			& \multirow{3}{*}{M=2} & accuracy & 0.908 & 0.572 & 0.803 & 0.811 & 0.986 & 0.982 & 0.988 & 0.966 \\
			& & f1 & 0.906 & 0.534 & 0.754 & 0.775 & 0.987 & 0.982 & 0.988 & 0.964 \\
			& & time & 0.028 & 0.086 & 0.038 & 0.021 & 0.034 & 0.105 & 0.038 & 0.021 \\ \bottomrule
			\multirow{9}{*}{clusters\_intersecting} & \multirow{3}{*}{M=0} & accuracy & 0.900 & 0.710 & 0.709 & 0.719 & 0.958 & 0.955 & 0.938 & 0.937 \\
			& & f1 & 0.897 & 0.689 & 0.690 & 0.698 & 0.958 & 0.955 & 0.938 & 0.936 \\
			& & time & 0.024 & 0.155 & 0.035 & 0.106 & 0.026 & 0.186 & 0.037 & 0.123 \\ \cmidrule(l){2-11}
			& \multirow{3}{*}{M=1} & accuracy & 0.904 & 0.694 & 0.715 & 0.722 & 0.967 & 0.963 & 0.948 & 0.948 \\
			& & f1 & 0.901 & 0.683 & 0.697 & 0.701 & 0.968 & 0.963 & 0.948 & 0.948 \\
			& & time & 0.027 & 0.090 & 0.036 & 0.020 & 0.030 & 0.103 & 0.037 & 0.020 \\ \cmidrule(l){2-11}
			& \multirow{3}{*}{M=2} & accuracy & 0.894 & 0.680 & 0.723 & 0.732 & 0.968 & 0.959 & 0.955 & 0.953 \\
			& & f1 & 0.892 & 0.690 & 0.700 & 0.704 & 0.968 & 0.959 & 0.955 & 0.953 \\
			& & time & 0.028 & 0.094 & 0.037 & 0.020 & 0.035 & 0.116 & 0.040 & 0.022 \\ \bottomrule
		\end{tabular}%
	}
\end{table}

\begin{table}[H]
	\centering
	\caption{Experiment 2 - Retrain with new class - SpatialRelations}
	\resizebox{.7\textwidth}{!}{%
		\begin{tabular}{@{}llllllllll@{}}
			\toprule
			& & \makecell{OvR\\LR} & \makecell{SCL\\LR} & \makecell{OvP\\LR} & \makecell{OvPSC\\LR} & \makecell{OvR\\SVM} & \makecell{SCL\\SVM} & \makecell{OvP\\SVM} & \makecell{OvPSC\\SVM} \\ \midrule
			\multirow{3}{*}{N2=devant} & accuracy & 0.789 & 0.789 & 0.789 & 0.789 & 0.804 & 0.804 & 0.804 & 0.804 \\
			& f1 & 0.780 & 0.780 & 0.780 & 0.780 & 0.795 & 0.795 & 0.795 & 0.795 \\
			& time & 0.028 & 0.028 & 0.011 & 0.017 & 0.377 & 0.378 & 0.183 & 0.189 \\ \midrule
			\multirow{3}{*}{N3=derriere} & accuracy & 0.703 & 0.703 & 0.706 & 0.706 & 0.737 & 0.737 & 0.722 & 0.722 \\
			& f1 & 0.700 & 0.700 & 0.703 & 0.703 & 0.735 & 0.735 & 0.718 & 0.718 \\
			& time & 0.056 & 0.181 & 0.024 & 0.092 & 1.081 & 1.356 & 0.306 & 0.454 \\ \midrule
			\multirow{3}{*}{N4=sous} & accuracy & 0.701 & 0.701 & 0.704 & 0.704 & 0.736 & 0.736 & 0.722 & 0.722 \\
			& f1 & 0.712 & 0.712 & 0.716 & 0.716 & 0.747 & 0.747 & 0.734 & 0.734 \\
			& time & 0.098 & 0.190 & 0.028 & 0.096 & 1.486 & 1.704 & 0.180 & 0.326 \\ \midrule
			\multirow{3}{*}{N5=sur} & accuracy & 0.710 & 0.708 & 0.710 & 0.711 & 0.740 & 0.741 & 0.728 & 0.728 \\
			& f1 & 0.734 & 0.732 & 0.735 & 0.736 & 0.760 & 0.761 & 0.753 & 0.752 \\
			& time & 0.162 & 0.234 & 0.027 & 0.118 & 1.891 & 1.937 & 0.147 & 0.308 \\ \midrule
			\multirow{3}{*}{N6=pres\_de} & accuracy & 0.653 & 0.653 & 0.653 & 0.654 & 0.682 & 0.682 & 0.671 & 0.670 \\
			& f1 & 0.609 & 0.609 & 0.593 & 0.595 & 0.621 & 0.621 & 0.608 & 0.607 \\
			& time & 0.190 & 0.297 & 0.010 & 0.110 & 2.800 & 2.990 & 0.190 & 0.384 \\ \midrule
			\multirow{3}{*}{N7=en\_face\_de} & accuracy & 0.616 & 0.615 & 0.615 & 0.618 & 0.642 & 0.608 & 0.634 & 0.632 \\
			& f1 & 0.525 & 0.530 & 0.503 & 0.509 & 0.539 & 0.516 & 0.521 & 0.516 \\
			& time & 0.235 & 0.209 & 0.018 & 0.110 & 3.549 & 2.446 & 0.381 & 0.445 \\ \midrule
			\multirow{3}{*}{N8=loin\_de} & accuracy & 0.596 & 0.592 & 0.601 & 0.602 & 0.622 & 0.554 & 0.614 & 0.613 \\
			& f1 & 0.503 & 0.499 & 0.483 & 0.483 & 0.511 & 0.461 & 0.479 & 0.483 \\
			& time & 0.297 & 0.183 & 0.016 & 0.095 & 3.957 & 1.582 & 0.156 & 0.260 \\ \midrule
			\multirow{3}{*}{N9=contre} & accuracy & 0.577 & 0.568 & 0.583 & 0.583 & 0.601 & 0.519 & 0.595 & 0.595 \\
			& f1 & 0.444 & 0.448 & 0.425 & 0.425 & 0.453 & 0.396 & 0.422 & 0.425 \\
			& time & 0.350 & 0.192 & 0.015 & 0.091 & 4.867 & 0.688 & 0.095 & 0.243 \\ \midrule
			\multirow{3}{*}{N10=au\_niveau\_de} & accuracy & 0.565 & 0.554 & 0.571 & 0.572 & 0.594 & 0.486 & 0.584 & 0.583 \\
			& f1 & 0.423 & 0.421 & 0.391 & 0.397 & 0.432 & 0.352 & 0.399 & 0.393 \\
			& time & 0.397 & 0.186 & 0.053 & 0.104 & 4.824 & 1.097 & 0.212 & 0.216 \\ \midrule
			\multirow{3}{*}{N11=au\_dessus\_de} & accuracy & 0.560 & 0.551 & 0.561 & 0.564 & 0.580 & 0.456 & 0.575 & 0.575 \\
			& f1 & 0.416 & 0.412 & 0.362 & 0.373 & 0.422 & 0.310 & 0.378 & 0.376 \\
			& time & 0.466 & 0.199 & 0.026 & 0.079 & 5.579 & 0.689 & 0.133 & 0.194 \\ \midrule
			\multirow{3}{*}{N12=dans} & accuracy & 0.557 & 0.551 & 0.561 & 0.562 & 0.583 & 0.453 & 0.574 & 0.573 \\
			& f1 & 0.430 & 0.427 & 0.386 & 0.382 & 0.437 & 0.291 & 0.394 & 0.388 \\
			& time & 0.542 & 0.211 & 0.042 & 0.058 & 5.223 & 0.940 & 0.058 & 0.098 \\ \midrule
			\multirow{3}{*}{N13=autour\_de} & accuracy & 0.557 & 0.550 & 0.561 & 0.562 & 0.581 & 0.446 & 0.572 & 0.571 \\
			& f1 & 0.452 & 0.445 & 0.409 & 0.404 & 0.450 & 0.278 & 0.399 & 0.394 \\
			& time & 0.592 & 0.164 & 0.036 & 0.057 & 5.339 & 1.086 & 0.052 & 0.088 \\ \midrule
			\multirow{3}{*}{N14=a\_l'exterieur\_de} & accuracy & 0.554 & 0.546 & 0.558 & 0.559 & 0.578 & 0.453 & 0.570 & 0.568 \\
			& f1 & 0.430 & 0.423 & 0.378 & 0.375 & 0.426 & 0.251 & 0.369 & 0.365 \\
			& time & 0.503 & 0.210 & 0.020 & 0.036 & 5.434 & 0.652 & 0.030 & 0.063 \\ \midrule
			\multirow{3}{*}{N15=aucun} & accuracy & 0.550 & 0.540 & 0.555 & 0.557 & 0.575 & 0.422 & 0.566 & 0.565 \\
			& f1 & 0.401 & 0.385 & 0.352 & 0.349 & 0.395 & 0.218 & 0.344 & 0.340 \\
			& time & 0.561 & 0.164 & 0.010 & 0.041 & 5.591 & 0.687 & 0.038 & 0.083 \\ \midrule
			\multirow{3}{*}{N16=le\_long\_de} & accuracy & 0.548 & 0.537 & 0.554 & 0.555 & 0.570 & 0.446 & 0.565 & 0.564 \\
			& f1 & 0.372 & 0.362 & 0.330 & 0.327 & 0.373 & 0.205 & 0.322 & 0.318 \\
			& time & 0.601 & 0.062 & 0.018 & 0.025 & 6.091 & 0.928 & 0.020 & 0.038 \\ \midrule
			\multirow{3}{*}{N17=par\_dela} & accuracy & 0.547 & 0.537 & 0.553 & 0.554 & 0.573 & 0.442 & 0.564 & 0.563 \\
			& f1 & 0.351 & 0.341 & 0.310 & 0.308 & 0.348 & 0.191 & 0.303 & 0.299 \\
			& time & 0.640 & 0.090 & 0.011 & 0.022 & 5.695 & 0.909 & 0.010 & 0.026 \\ \bottomrule
		\end{tabular}%
	}
\end{table}

\begin{table}[H]
	\centering
	\caption{Experiment 2 - Retrain with new class - clusters\_veryFar}
	\resizebox{\textwidth}{!}{%
		\begin{tabular}{@{}llllllllll@{}}
			\toprule
			& & \makecell{OvR\\LR} & \makecell{SCL\\LR} & \makecell{OvP\\LR} & \makecell{OvPSC\\LR} & \makecell{OvR\\SVM} & \makecell{SCL\\SVM} & \makecell{OvP\\SVM} & \makecell{OvPSC\\SVM} \\ \midrule
			\multirow{3}{*}{N2=7} & accuracy & 1.000 & 1.000 & 1.000 & 1.000 & 1.000 & 1.000 & 1.000 & 1.000 \\
			& f1 & 1.000 & 1.000 & 1.000 & 1.000 & 1.000 & 1.000 & 1.000 & 1.000 \\
			& time & 0.005 & 0.007 & 0.002 & 0.005 & 0.007 & 0.009 & 0.002 & 0.005 \\ \midrule
			\multirow{3}{*}{N3=1} & accuracy & 1.000 & 1.000 & 1.000 & 1.000 & 1.000 & 1.000 & 1.000 & 1.000 \\
			& f1 & 1.000 & 1.000 & 1.000 & 1.000 & 1.000 & 1.000 & 1.000 & 1.000 \\
			& time & 0.008 & 0.018 & 0.002 & 0.010 & 0.011 & 0.021 & 0.002 & 0.012 \\ \midrule
			\multirow{3}{*}{N4=4} & accuracy & 1.000 & 1.000 & 0.753 & 1.000 & 1.000 & 1.000 & 1.000 & 1.000 \\
			& f1 & 1.000 & 1.000 & 0.668 & 1.000 & 1.000 & 1.000 & 1.000 & 1.000 \\
			& time & 0.011 & 0.024 & 0.002 & 0.016 & 0.015 & 0.030 & 0.003 & 0.019 \\ \midrule
			\multirow{3}{*}{N5=8} & accuracy & 1.000 & 0.947 & 0.802 & 1.000 & 1.000 & 1.000 & 1.000 & 1.000 \\
			& f1 & 1.000 & 0.943 & 0.734 & 1.000 & 1.000 & 1.000 & 1.000 & 1.000 \\
			& time & 0.014 & 0.031 & 0.002 & 0.021 & 0.021 & 0.038 & 0.003 & 0.026 \\ \midrule
			\multirow{3}{*}{N6=0} & accuracy & 1.000 & 0.668 & 0.809 & 1.000 & 1.000 & 1.000 & 1.000 & 1.000 \\
			& f1 & 1.000 & 0.584 & 0.753 & 1.000 & 1.000 & 1.000 & 1.000 & 1.000 \\
			& time & 0.018 & 0.038 & 0.002 & 0.026 & 0.026 & 0.041 & 0.003 & 0.031 \\ \midrule
			\multirow{3}{*}{N7=5} & accuracy & 1.000 & 0.568 & 0.836 & 1.000 & 1.000 & 1.000 & 1.000 & 1.000 \\
			& f1 & 1.000 & 0.485 & 0.788 & 1.000 & 1.000 & 1.000 & 1.000 & 1.000 \\
			& time & 0.021 & 0.039 & 0.002 & 0.031 & 0.033 & 0.057 & 0.004 & 0.039 \\ \midrule
			\multirow{3}{*}{N8=6} & accuracy & 1.000 & 0.492 & 0.856 & 1.000 & 1.000 & 1.000 & 1.000 & 1.000 \\
			& f1 & 1.000 & 0.424 & 0.815 & 1.000 & 1.000 & 1.000 & 1.000 & 1.000 \\
			& time & 0.025 & 0.046 & 0.002 & 0.037 & 0.041 & 0.058 & 0.003 & 0.045 \\ \midrule
			\multirow{3}{*}{N9=3} & accuracy & 1.000 & 0.545 & 0.766 & 1.000 & 1.000 & 0.869 & 1.000 & 1.000 \\
			& f1 & 1.000 & 0.449 & 0.693 & 1.000 & 1.000 & 0.820 & 1.000 & 1.000 \\
			& time & 0.030 & 0.047 & 0.002 & 0.040 & 0.048 & 0.059 & 0.003 & 0.050 \\ \bottomrule
		\end{tabular}%
	}
\end{table}

\begin{table}[H]
	\centering
	\caption{Experiment 2 - Retrain with new class - clusters\_far}
	\resizebox{\textwidth}{!}{%
		\begin{tabular}{@{}llllllllll@{}}
			\toprule
			& & \makecell{OvR\\LR} & \makecell{SCL\\LR} & \makecell{OvP\\LR} & \makecell{OvPSC\\LR} & \makecell{OvR\\SVM} & \makecell{SCL\\SVM} & \makecell{OvP\\SVM} & \makecell{OvPSC\\SVM} \\ \midrule
			\multirow{3}{*}{N2=1} & accuracy & 0.988 & 0.988 & 0.988 & 0.988 & 0.606 & 0.606 & 0.606 & 0.606 \\
			& f1 & 0.988 & 0.988 & 0.988 & 0.988 & 0.469 & 0.469 & 0.469 & 0.469 \\
			& time & 0.005 & 0.007 & 0.002 & 0.005 & 0.006 & 0.008 & 0.002 & 0.005 \\ \midrule
			\multirow{3}{*}{N3=7} & accuracy & 1.000 & 0.919 & 0.992 & 0.992 & 1.000 & 0.988 & 0.736 & 0.472 \\
			& f1 & 1.000 & 0.918 & 0.992 & 0.992 & 1.000 & 0.988 & 0.646 & 0.335 \\
			& time & 0.008 & 0.018 & 0.002 & 0.010 & 0.010 & 0.021 & 0.002 & 0.012 \\ \midrule
			\multirow{3}{*}{N4=4} & accuracy & 0.997 & 1.000 & 0.917 & 0.917 & 1.000 & 0.805 & 0.802 & 0.603 \\
			& f1 & 0.997 & 1.000 & 0.915 & 0.914 & 1.000 & 0.783 & 0.735 & 0.469 \\
			& time & 0.011 & 0.025 & 0.002 & 0.016 & 0.015 & 0.030 & 0.002 & 0.019 \\ \midrule
			\multirow{3}{*}{N5=8} & accuracy & 1.000 & 1.000 & 0.930 & 0.923 & 1.000 & 0.538 & 0.838 & 0.680 \\
			& f1 & 1.000 & 1.000 & 0.929 & 0.922 & 1.000 & 0.451 & 0.785 & 0.563 \\
			& time & 0.014 & 0.030 & 0.002 & 0.021 & 0.018 & 0.035 & 0.003 & 0.025 \\ \midrule
			\multirow{3}{*}{N6=0} & accuracy & 0.950 & 0.929 & 0.942 & 0.935 & 1.000 & 0.669 & 0.863 & 0.731 \\
			& f1 & 0.950 & 0.928 & 0.941 & 0.935 & 1.000 & 0.587 & 0.821 & 0.632 \\
			& time & 0.018 & 0.035 & 0.002 & 0.026 & 0.022 & 0.041 & 0.002 & 0.031 \\ \midrule
			\multirow{3}{*}{N7=6} & accuracy & 0.978 & 0.894 & 0.948 & 0.944 & 1.000 & 0.624 & 0.882 & 0.767 \\
			& f1 & 0.978 & 0.891 & 0.948 & 0.944 & 1.000 & 0.531 & 0.847 & 0.680 \\
			& time & 0.022 & 0.041 & 0.002 & 0.030 & 0.025 & 0.047 & 0.003 & 0.037 \\ \midrule
			\multirow{3}{*}{N8=5} & accuracy & 0.982 & 0.726 & 0.954 & 0.951 & 1.000 & 0.608 & 0.896 & 0.795 \\
			& f1 & 0.983 & 0.732 & 0.954 & 0.951 & 1.000 & 0.507 & 0.866 & 0.720 \\
			& time & 0.025 & 0.047 & 0.002 & 0.035 & 0.029 & 0.054 & 0.002 & 0.043 \\ \midrule
			\multirow{3}{*}{N9=2} & accuracy & 0.981 & 0.689 & 0.914 & 0.921 & 0.997 & 0.588 & 0.906 & 0.816 \\
			& f1 & 0.982 & 0.691 & 0.910 & 0.918 & 0.997 & 0.488 & 0.881 & 0.746 \\
			& time & 0.029 & 0.048 & 0.002 & 0.038 & 0.038 & 0.057 & 0.003 & 0.047 \\ \bottomrule
		\end{tabular}%
	}
\end{table}

\begin{table}[H]
	\centering
	\caption{Experiment 2 - Retrain with new class - clusters\_close}
	\resizebox{\textwidth}{!}{%
		\begin{tabular}{@{}llllllllll@{}}
			\toprule
			& & \makecell{OvR\\LR} & \makecell{SCL\\LR} & \makecell{OvP\\LR} & \makecell{OvPSC\\LR} & \makecell{OvR\\SVM} & \makecell{SCL\\SVM} & \makecell{OvP\\SVM} & \makecell{OvPSC\\SVM} \\ \midrule
			\multirow{3}{*}{N2=7} & accuracy & 0.964 & 0.964 & 0.964 & 0.964 & 0.600 & 0.600 & 0.600 & 0.600 \\
			& f1 & 0.963 & 0.963 & 0.963 & 0.963 & 0.467 & 0.467 & 0.467 & 0.467 \\
			& time & 0.005 & 0.007 & 0.002 & 0.005 & 0.007 & 0.008 & 0.002 & 0.005 \\ \midrule
			\multirow{3}{*}{N3=3} & accuracy & 0.984 & 0.984 & 0.936 & 0.936 & 0.988 & 0.770 & 0.719 & 0.469 \\
			& f1 & 0.984 & 0.984 & 0.936 & 0.936 & 0.988 & 0.715 & 0.636 & 0.334 \\
			& time & 0.008 & 0.020 & 0.002 & 0.011 & 0.011 & 0.022 & 0.002 & 0.013 \\ \midrule
			\multirow{3}{*}{N4=1} & accuracy & 0.958 & 0.901 & 0.931 & 0.931 & 0.985 & 0.817 & 0.787 & 0.405 \\
			& f1 & 0.957 & 0.894 & 0.931 & 0.931 & 0.985 & 0.798 & 0.727 & 0.281 \\
			& time & 0.011 & 0.023 & 0.002 & 0.016 & 0.013 & 0.027 & 0.002 & 0.020 \\ \midrule
			\multirow{3}{*}{N5=2} & accuracy & 0.937 & 0.891 & 0.928 & 0.928 & 0.986 & 0.810 & 0.828 & 0.365 \\
			& f1 & 0.936 & 0.888 & 0.927 & 0.927 & 0.986 & 0.762 & 0.781 & 0.255 \\
			& time & 0.015 & 0.031 & 0.002 & 0.022 & 0.016 & 0.036 & 0.003 & 0.027 \\ \midrule
			\multirow{3}{*}{N6=0} & accuracy & 0.965 & 0.931 & 0.923 & 0.923 & 0.982 & 0.858 & 0.851 & 0.465 \\
			& f1 & 0.965 & 0.931 & 0.923 & 0.923 & 0.982 & 0.830 & 0.814 & 0.334 \\
			& time & 0.018 & 0.037 & 0.002 & 0.026 & 0.021 & 0.044 & 0.003 & 0.032 \\ \midrule
			\multirow{3}{*}{N7=8} & accuracy & 0.949 & 0.904 & 0.900 & 0.919 & 0.977 & 0.806 & 0.866 & 0.538 \\
			& f1 & 0.949 & 0.903 & 0.899 & 0.919 & 0.978 & 0.748 & 0.835 & 0.415 \\
			& time & 0.022 & 0.041 & 0.002 & 0.031 & 0.026 & 0.048 & 0.003 & 0.039 \\ \midrule
			\multirow{3}{*}{N8=4} & accuracy & 0.941 & 0.662 & 0.793 & 0.810 & 0.989 & 0.663 & 0.878 & 0.592 \\
			& f1 & 0.940 & 0.632 & 0.745 & 0.762 & 0.989 & 0.598 & 0.851 & 0.473 \\
			& time & 0.025 & 0.044 & 0.002 & 0.036 & 0.028 & 0.056 & 0.003 & 0.045 \\ \midrule
			\multirow{3}{*}{N9=5} & accuracy & 0.913 & 0.634 & 0.767 & 0.782 & 0.992 & 0.602 & 0.888 & 0.630 \\
			& f1 & 0.911 & 0.599 & 0.722 & 0.739 & 0.992 & 0.535 & 0.864 & 0.517 \\
			& time & 0.029 & 0.054 & 0.002 & 0.041 & 0.033 & 0.061 & 0.003 & 0.051 \\ \bottomrule
		\end{tabular}%
	}
\end{table}

\begin{table}[H]
	\centering
	\caption{Experiment 2 - Retrain with new class - clusters\_intersecting}
	\resizebox{\textwidth}{!}{%
		\begin{tabular}{@{}llllllllll@{}}
			\toprule
			& & \makecell{OvR\\LR} & \makecell{SCL\\LR} & \makecell{OvP\\LR} & \makecell{OvPSC\\LR} & \makecell{OvR\\SVM} & \makecell{SCL\\SVM} & \makecell{OvP\\SVM} & \makecell{OvPSC\\SVM} \\ \midrule
			\multirow{3}{*}{N2=7} & accuracy & 1.000 & 1.000 & 1.000 & 1.000 & 1.000 & 1.000 & 1.000 & 1.000 \\
			& f1 & 1.000 & 1.000 & 1.000 & 1.000 & 1.000 & 1.000 & 1.000 & 1.000 \\
			& time & 0.006 & 0.007 & 0.002 & 0.005 & 0.006 & 0.008 & 0.002 & 0.005 \\ \midrule
			\multirow{3}{*}{N3=8} & accuracy & 0.996 & 0.954 & 0.983 & 0.983 & 0.996 & 0.992 & 0.988 & 0.988 \\
			& f1 & 0.996 & 0.954 & 0.983 & 0.983 & 0.996 & 0.992 & 0.987 & 0.987 \\
			& time & 0.008 & 0.019 & 0.002 & 0.011 & 0.010 & 0.022 & 0.002 & 0.012 \\ \midrule
			\multirow{3}{*}{N4=3} & accuracy & 0.987 & 0.934 & 0.840 & 0.834 & 0.975 & 0.972 & 0.981 & 0.981 \\
			& f1 & 0.987 & 0.934 & 0.829 & 0.818 & 0.975 & 0.972 & 0.981 & 0.981 \\
			& time & 0.011 & 0.026 & 0.002 & 0.016 & 0.012 & 0.030 & 0.002 & 0.019 \\ \midrule
			\multirow{3}{*}{N5=4} & accuracy & 0.962 & 0.912 & 0.663 & 0.668 & 0.970 & 0.952 & 0.967 & 0.967 \\
			& f1 & 0.962 & 0.913 & 0.592 & 0.584 & 0.970 & 0.952 & 0.967 & 0.967 \\
			& time & 0.014 & 0.031 & 0.002 & 0.021 & 0.016 & 0.038 & 0.002 & 0.026 \\ \midrule
			\multirow{3}{*}{N6=5} & accuracy & 0.941 & 0.870 & 0.688 & 0.688 & 0.962 & 0.954 & 0.954 & 0.954 \\
			& f1 & 0.940 & 0.875 & 0.631 & 0.621 & 0.962 & 0.954 & 0.954 & 0.954 \\
			& time & 0.018 & 0.037 & 0.002 & 0.026 & 0.021 & 0.043 & 0.002 & 0.032 \\ \midrule
			\multirow{3}{*}{N7=2} & accuracy & 0.955 & 0.848 & 0.654 & 0.645 & 0.951 & 0.960 & 0.950 & 0.950 \\
			& f1 & 0.954 & 0.861 & 0.607 & 0.591 & 0.950 & 0.960 & 0.950 & 0.950 \\
			& time & 0.021 & 0.041 & 0.002 & 0.032 & 0.028 & 0.051 & 0.002 & 0.039 \\ \midrule
			\multirow{3}{*}{N8=0} & accuracy & 0.899 & 0.764 & 0.685 & 0.676 & 0.968 & 0.962 & 0.948 & 0.948 \\
			& f1 & 0.897 & 0.755 & 0.646 & 0.631 & 0.968 & 0.962 & 0.948 & 0.948 \\
			& time & 0.025 & 0.049 & 0.002 & 0.036 & 0.031 & 0.056 & 0.003 & 0.044 \\ \midrule
			\multirow{3}{*}{N9=6} & accuracy & 0.894 & 0.663 & 0.708 & 0.704 & 0.970 & 0.963 & 0.949 & 0.949 \\
			& f1 & 0.892 & 0.634 & 0.675 & 0.666 & 0.970 & 0.963 & 0.949 & 0.949 \\
			& time & 0.029 & 0.052 & 0.003 & 0.041 & 0.037 & 0.064 & 0.003 & 0.050 \\ \bottomrule
		\end{tabular}%
	}
\end{table}

\end{document}